# Balanced background and explanation data are needed in explaining deep learning models with SHAP: An empirical study on clinical decision making

Mingxuan Liu[1], Yilin Ning[1], Han Yuan[1], Marcus Eng Hock Ong[2,3], Nan Liu[1,2,4,5]

[1] Centre for Quantitative Medicine, Duke-NUS Medical School, Singapore

[2] Programme in Health Services and Systems Research, Duke-NUS Medical School, Singapore

[3] Department of Emergency Medicine, Singapore General Hospital, Singapore

[4] SingHealth AI Health Program, Singapore Health Services, Singapore

[5] Institute of Data Science, National University of Singapore, Singapore

**Abstract**

**Objective:** Shapley additive explanations (SHAP) is a popular post-hoc technique for explaining black box models. While the impact of data imbalance on predictive models has been extensively studied, it remains largely unknown with respect to SHAP-based model explanations. This study sought to investigate the effects of data imbalance on SHAP explanations for deep learning models, and to propose a strategy to mitigate these effects.

**Materials and Methods**: We propose to adjust class distributions in the background and explanation data in SHAP when explaining black box models. Our data balancing strategy is to compose background data and explanation data with an equal distribution of classes. To evaluate the effects of data adjustment on model explanation, we propose to use the beeswarm plot as a qualitative tool to identify "abnormal" explanation artifacts, and quantitatively test the consistency between variable importance and prediction power. We demonstrated our proposed approach in an empirical study that predicted inpatient mortality using the Medical Information Mart for Intensive Care (MIMIC-III) data and a multilayer perceptron.

**Results**: Using the data balancing strategy would allow us to reduce the number of the artifacts in the beeswarm plot, thus mitigating the negative effects of data imbalance. Additionally, with the balancing strategy, the top-ranked variables from the corresponding importance ranking demonstrated improved discrimination power.

**Discussion and Conclusion**: Our findings suggest that balanced background and explanation data could help reduce the noise in explanation results induced by skewed data distribution and improve the reliability of variable importance ranking. Furthermore, these balancing procedures improve the potential of SHAP in identifying patients with abnormal characteristics in clinical applications.

# 1 Background and Significance

As deep learning models become increasingly popular, interpretability remains a barrier to their adoption, as these black boxes maintain an opaque relationship between input and output. Lack of interpretability constrains models' acceptance among users and raises legal and ethical concerns, especially in the clinical field[1]. To improve the interpretability of deep learning models, inherently interpretable models can be developed using techniques such as knowledge distillation[2], which often involves trade-offs between predictivity and interpretability[3]. In contrast, post hoc explanations provide a certain degree of interpretability to deep learning models without impairing their high predictive accuracy.

Several popular post hoc explanation methods are available for deep learning models. The Shapley additive explanations (SHAP)[4] family is widely used. Based on the Shapley value in game theory[5], SHAP is closely related to other post hoc methods such as Local Interpretable Model-agnostic Explanations (LIME)[6], Deep Learning Important FeaTures (DeepLIFT)[7,8] and integrates gradients[9]. SHAP quantifies variable attributions for each observation, yielding local explanations that measure the influence of each variable on the prediction outcome. The global variable importance[10], i.e., the overall contribution of each variable to the model, can be derived from the mean absolute SHAP values across observations[4].

Model explanations using SHAP have been widely adopted[11-17]. SHAP values provide users with practical information, such as which variables should receive more attention in clinical practice. Additionally, important variables may be prioritized for the construction of lite models, with SHAP values serving as a variable selection method for high-dimensional inputs[16]. As a local explanation method, SHAP produces variable contributions exclusively for the observations supplied by users. Ideally, these observations, which are known as the explanation data, should be representative of all possible predictor profiles, allowing the explanation results to express comprehensively how variables might influence predictions. Similarly, the background data required for SHAP value calculation should also provide comprehensive prior information for model reference and replacement of missing values[4]. According to our knowledge, existing research on SHAP has paid little attention to the background and explanation data. A simple feed of the entire dataset or a random subset may be insufficient to deal with unbalanced datasets, e.g., medical datasets that are often intrinsically unbalanced due to rare events[18, 19].

Although there is evidence that data imbalance impairs model prediction performance[20], and many solutions have been proposed[21-23], its effects on model explanations via SHAP have largely been neglected. Data imbalance is typically characterized by highly unequal distributions among its classes[24]. With a binary outcome, data imbalance refers to a situation where the majority class (i.e., the predominant class of outcome) substantially outnumbers the minority class. This imbalance raises the concern that noise may be introduced into explanations due to the majority's dominance and the extremely limited information about the minority class, thus impairing interpretation. In SHAP applications, we hypothesize that unbalanced background and explanation data will lead to explanation bias, which will negatively impact variable contributions – a critical consideration in many clinical studies[1, 13-15]. Additionally, we anticipate that by balancing background data and manipulating explanation data appropriately, we may improve SHAP explanations and distinguish meaningful signals from noise.

In this study, we examine the effects of background data and explanation data on SHAP-based explanations for deep learning models, and demonstrate our proposed solutions using the Medical Information Mart for Intensive Care (MIMIC-III)[25] database. This study seeks to provide empirical evidence regarding the impact of data imbalance on SHAP-based interpretability, which could help guide future choices of background and explanation data and offer a practical supplement for SHAP applications.

## 2 Materials and Methods

In this section, we describe how SHAP explains deep learning models and present a balancing strategy for background and explanation data. Thereafter, we elaborate the process for evaluating the effects of data adjustments. Lastly, we demonstrate how this data balancing strategy can be implemented in practice using the MIMIC-III data.

We follow the common practice when developing and explaining a deep learning model, where we apply the hold-out method to divide the data into non-overlapping training set, validation set, and test set, stratified by the outcome. Since this study focuses on model explanation, we assume the deep learning model has been trained using the training set and fine-tuned using the validation set.

**2.1 Explaining deep learning models with SHAP**

SHAP considers the attribution of variable effects to be confounded rather than isolated[1], hence accounting for the interaction between variables. Predictions of black box models are explained as a linear combination of individual variables' contributions, which is easily understood and validated by users who may not have expertise in machine learning.

SHAP employs a variety of algorithms for deep learning models to measure variable contributions. Kernel SHAP is a model-agnostic approximation that is not tied to any specific type of model. Gradient SHAP supports differentiable models including deep learning models and some machine learning models like support vector machine. Deep SHAP, adapted from DeepLIFT algorithm[7], is another popular explainer, which is specifically designed for deep learning models. This study will utilize Deep SHAP for demonstration since it is efficient in approximation and tailored for deep learning models.

For computing SHAP values, researchers usually specify a subset of the entire dataset as the background data to approximate the expected values of the majority and minority classes. By summing variables' contributions (i.e., SHAP values) for an observation, we can determine the deviation of the model's prediction from the average (expected) value. Essentially, the background data provides the SHAP explainer with the prior information about the population, thereby influencing the final SHAP values. With the dominance of the majority class, an unbalanced background data will lead to a much lower baseline for observations in the minority class, and probably lead to overestimate their SHAP values, as SHAP describes the differences between the prediction and average value[4].

When a SHAP explainer is constructed with the model and background data, it will be applied to the explanation data and used to explain user-specified observations where the corresponding SHAP value for each variable can be derived. The choice of explanation data directly affects both the local explanations and the global importance ranking. When explanation data are unbalanced, the explanation results may be dominated by the majority class, obscuring the explanations from the minority class.

## 2.2 Balancing the background and explanation data

In the case of a black box model, the negative effects of data imbalance on model explanations may be offset by appropriate background and explanation data in SHAP. The minority-overall rate, defined as the ratio of minority cases to all cases, serves as a measure of data imbalance. Several strategies have been considered to sample background data, including the use of the entire test set[14], the entire training set[16] or external balanced data[12, 17] instead of the existing training, validation or test sets. For the explanation data, under-sampled balanced validation set[15] and the whole training set[13] have been explored. We sample background data from the training set that contained the most population in the cohort in order to provide comprehensive prior information for the SHAP explainer, and compose explanation data from the validation set to guarantee no overlap with background data. The test set is preserved for evaluating the SHAP explanations after these adjustments.

The first step is to deal with data imbalance in the background data, since it provides prior information on variable distribution that is critical to calculating SHAP values. To generate a set of background data with a fixed size, *N*, from the training set, a simple random sampling yields the same minority-overall rate $p_0$ as in the original data. We propose to balance the background data to a higher minority-overall rate of $p_0 < p \leq 0.5$, through a random selection of *Np* observations from the minority class and *N(1-p)* observations from the majority class. As *p* increases, more attention will be drawn to the minority class when explaining the fitted model. Previous studies[12, 17] suggest that $p = 0.5$ is an appropriate choice.

Next, we apply under-sampling[26-28] to the validation set to compose the explanation data, in which we select all samples from the minority class, but only a fraction of the majority class. In this way, the minority-overall rate is increased to the same level as in the background data, ensuring that the distribution conforms to the prior information. While over-sampling (i.e., repeated sampling or synthesis of observations for the minority class) is a simple yet effective data balancing strategy, under-sampling is generally preferred in the medical field for two reasons: (1) explanation results would be more persuasive without synthetic data for the explanation, and (2) the reduction in the number of explanation data could save significant computation time. One potential limitation of under-sampling is the possibility of information loss in the majority class[26]. We address this problem by applying a K-means-based under-sampling approach, in which we perform clustering analysis on the observations in the majority class to capture group patterns. The number of cases to be randomly selected from each cluster will be equal, and for clusters with insufficient patients, all cases will be selected. We use the elbow method to determine the number of clusters, where we simultaneously minimize within-cluster variability and the number of clusters[29].

**2.3 Evaluating the effects of adjustments to the background and explanation Data**

For each combination of background and explanation data, we can generate corresponding SHAP values. Then we evaluate the effects in two ways: (1) qualitatively examining the beeswarm plot from SHAP, and (2) quantitatively assessing the predictive ability of models that use only the top few most important variables selected by SHAP. These two aspects of evaluation are consistent with the two dimensions of variable importance – explanatory and predictive[30].

*2.3.1 Qualitative analyses*

For qualitative validation, we use the beeswarm plot, which shows both global and local explanations. As visualized in the plot, the color of the dots and the SHAP values indicated by the x-axis can illustrate the association between the variable's value (e.g., age by year) and contribution (e.g., age's corresponding SHAP value). The spread of the dots displays the variance of the variable's overall contribution, and the distribution of the data points can be examined to detect abnormal patterns. In the y-axis, the order of variable importance is determined by the mean of absolute SHAP values over all explanation observations as a heuristic measure of global variable importance. The beeswarm plot may be easily generated using the SHAP Python package[4].

*2.3.2 Quantitative analyses*

Qualitative analyses focus on the beeswarm plot, which is critical for SHAP applications. Quantitatively, we focused on the importance ranking, where the variable importance is generated by the mean absolute variable contributions across the observations in the explanation data. We tested the predictive ability of top $k$ ($k$ = 3, 4, …20) variables, assuming that top $k$ variables from a good ranking would have higher predictive power than top $k$ variables from a poor ranking. For a specific $k$, we build models using the top $k$ variables from the importance rankings before and after adjusting the background and explanation data. These models are in the same architecture and with the same hyper-parameter selection strategy, and the only difference is the input, i.e., top-$k$ variables from different importance rankings. Then we compare their discriminative ability in terms of AUC, using the bootstrapping method to compute 95% confidence intervals (CI).

**2.4 Experiments**

We conducted an empirical study using the MIMIC-III database, which is a publicly available critical care dataset obtained from the intensive care units (ICUs) of the Beth Israel Deaconess Medical Center (BIDMC) between 2001 and 2012. It contains de-identified information on 44,918 patients.[25] The ratio used in dataset splitting was 7:1:2 for composing training, validation and test sets. Twenty-one continuous variables were used for analysis, including patient demographics, vital signs, and laboratory tests. The outcome was binary (death or survival), with a mortality rate of 8.8%. In this manner, the dataset was unbalanced, with patients who died forming the minority class and patients who survived constituting the majority class.

After constructing the deep learning model that will be explained, i.e., multilayer perceptron (MLP) in this study, we first created background datasets of fixed size with varying minority-overall rate ($p$) of 0.088 and 0.5. The minority-overall rate of 0.088 (i.e., the original event rate) was considered a reference point. By increasing $p$ to 0.5, the majority and minority classes held almost equal baselines for the computation of SHAP values. Following that, for each background, we examined two versions of the explanation data: (1) the original validation set with a minority-overall rate equal to the event rate of 0.088, and (2) an under-sampled validation set with the same minority-overall rate as the background data.

To enable convenient application, we provide a Python library named "BalanceSHAP"[31]. Through this library, the procedures of composing background and explanation data can be easily implemented. It also inherits computation and plotting methods from the original SHAP library[4], providing a convenient interface for SHAP calculation, especially in the scenario of data imbalance.

# 3 Results

This section describes the model implementation and reports the qualitative and quantitative evaluation results for adjustments made to the background and explanation data in SHAP.

## 3.1 Model implementation

We constructed an explanation model comprising of four linear layers activated by rectified linear units (ReLU), two dropout layers, and an output layer with a Sigmoid activation function. We optimized the model by minimizing the cross-entropy loss, using weights of [0.92, 0.08] for the minority (mortality) and majority (survival) classes, respectively. The learning rate was set to 0.01 in the experiment of 200 epochs. On the test set, this model achieved an AUC of 0.827 (95% CI: 0.814 – 0.841).

## 3.2 Qualitative evaluation results

We begin with examining the impact of the adjustments to background data on SHAP values, where the original validation set was used as the explanation data. Next, we assessed the additional impact of balancing explanation data on model explanations.

### *3.2.1 SHAP with original and balanced background dataset*

Using the original validation set as explanation data, Figure 1 compares the beeswarm plots generated based on unbalanced (with a minority-overall rate of 0.088; see Figure 1a) and balanced background data (with minority-overall rates of 0.5; see Figure 1b). While we observed variations in the variable importance ranking, the most and least important variables remained relatively constant, such as blood urea nitrogen (BUN), age, and oxygen saturation ($SpO_2$).

There were several abnormal points (see points *a-n* in Figure 1b) in the SHAP values based on the unbalanced background data that were inconsistent with the overall association between variable value and its contribution towards the prediction, i.e., SHAP value. As an example, point *b* is an "abnormal point" when interpreting the importance of age, as it is inconsistent with the general trend where old age contributed positively to mortality. This abnormality may arise due to two reasons: genuine outlying observation for particular patients that warrant further investigation, or inaccurate SHAP values that require correction before interpretation. Generally, the abnormal points *a-n* (except points *i*, *m*, and *l*) were removed after balancing the background data (see Figure 1b), thereby improving the understanding of the association and interpretation. Nevertheless, some abnormal points (for example, points *i*, *m*, and *l* in Figure 1) remained, which could be addressed in the subsequent adjustments made to the explanation data.

### *3.2.2 SHAP with original and balanced explanation datasets*

Subsequently, we examined the benefits of adjusting the explanation data to match the minority-overall rate in the background data, such that the data to be explained is consistent with the prior information provided to SHAP. For the K-means-based under-sampling, the number of clusters was determined as three by the elbow method[32]. Using the balanced background data, we compared the beeswarm plots generated by unbalanced (original validation set, with a minority-overall rate of 0.088; see Figure 2a) and balanced (under-sampled, with a minority-overall rate of 0.5; see Figure 2b) explanation data. Apart from the effects of the balanced background, Figure 2b displays the trend patterns more clearly for potassium and diastolic blood pressure. As compared to Figure 2a, points *i* and *m* have been corrected; however, point *l* (for chloride) remains unchanged.

An examination of chloride further revealed that point *l* had the fourth-highest chloride level (0.723, top 0.25%), with the corresponding SHAP value significantly greater than other observations with top-10 chloride levels. In addition, for point *l*, when we replaced its chloride level with that of the most similar observation among the explanation data (0.571, top 22.5%), while retaining all other variable values, the SHAP value for chloride became negative, consistent with other observations with high chloride values. These results suggest that point *l* likely represents a true outlier in the data that was identified through visualization only after removing the noise induced by data imbalance.

### 3.3 Quantitative evaluation results

The qualitative evaluation revealed local changes in SHAP values after adjusting the background and explanation data. We then quantitatively evaluated the impact of corresponding global changes, i.e., changes in variable importance ranking, by assessing the predictive performance of models constructed with top-ranked variables. Figure 3 displays the AUC values for MLP models built with top $k$ ($k$ = 3, 4, …, 20) variables which were extracted from SHAP-based importance rankings. These importance rankings were obtained using background datasets with minority-overall rates of 0.088 and 0.5, respectively, with and without under-sampling the explanation data to the same minority-overall rate. The minority-overall rate of 0.088, which was the event rate in the original data, served as the baseline for comparison.

When both background and explanation data were balanced, we found that MLPs constructed with top-ranked variables generally outperformed models built based on the original unbalanced data. Additionally, the program running time could be significantly reduced with the balanced explanation data, as compared to the unbalanced version (3.72 minutes vs 17.07 minutes, run on the Windows 11 system with Intel Core i5 Processor with 4 cores and 1.60GHz, 16GB RAM), since the computation time increases linearly with the sample size of the explanation data (sample size 1,579 vs 8,982).

## 4 Discussion

This study highlighted the impact of data imbalance (a common phenomenon in clinical applications) on explaining black box models via SHAP and proposed a balancing strategy for background and explanation data to offset the impact, which was qualitatively and quantitatively evaluated. The balancing strategy can eliminate ambiguous information resulting from data imbalance and provide a more precise explanation, leading to trustworthy clinical insights in decision-making. Moreover, this balancing strategy enables researchers to develop parsimonious models with reasonable performance and improved interpretability by choosing variables with high predictability from the importance ranking. To aid the implementation of the balancing strategy and calculation of SHAP values, we have developed the BalanceSHAP Python library[31] to assist researchers using SHAP in the case of data imbalance.

Data imbalance can result in inaccurate SHAP explanations, which appears as abnormal points on the beeswarm plot that deviate significantly from the general trend of association between variable values and corresponding SHAP values. The trends in the beeswarm plot have been found useful for interpreting prediction models in clinical settings, with occasional abnormal points being ignored[1, 13, 33]. When multiple abnormal points are observed for a variable, especially when they present strong impacts on the outcome (e.g. points *a-g* in Figure 1), it becomes difficult to interpret the overall influence of this variable on the outcome or to anticipate the outcome of a clinical intervention based on this variable[1, 33]. As demonstrated in our empirical study, balancing the background and explanation data reduced abnormal points caused by data imbalance, hence reducing confusion and incorrect interpretation.

Apart from erroneous abnormal points that arise from inaccurate SHAP values due to imbalance and hinders decision-making, some abnormal points may indicate genuine outliers in the dataset that provide useful information for further analysis[34]. Since the background and explanation data only influence the computation of SHAP values[4], their adjustments can minimize erroneous abnormal points but has little impact on the genuine ones, e.g. point *l* that persisted in Figures 1 and 2 despite various data adjustments. By relieving confusion caused by the first type of abnormal points, this balancing strategy may prove useful in identifying patients with concerning conditions[35] (i.e. the second type) in clinical applications, allowing for more personalized interventions.

Another benefit of our proposed balancing strategy is the reduction in the size of explanation data and hence computation time. Using a large dataset (such as the entire training set[13]) as the explanation data may seem logical, but it can be extremely time-consuming[4]. This study proposed to apply the widely used K-means-based under-sampling[26-28] to balance the smaller validation set and capture the representative patterns in the majority class, and then use it to generate local explanations. In this manner, the time-consuming problem is mitigated without adversely affecting the explanation results or the predictive ability of the top-ranked variables. By using the K-means-based method to capture the profiles of the majority class, we can control the size of the explanation data without losing information for explanations.

As a result of our proposed balancing procedures, the SHAP-based variable ranking is more aligned with variable contribution to predictions, as reflected in the improved predictive performance of models based on the top few variables. In this way, this balancing strategy enhances the variable importance assessment in the dimension of predictivity[36]. Selecting the top variables in the post-balanced importance ranking allows researchers to develop well-performing predictive models with fewer predictors, which is especially valuable in clinical studies that favor simple models over complex ones to enhance interpretability[37, 38].

Except for the balanced background and explanation data, we also used those with minority-overall rate $p < 0.5$. According to eTable 1, we observed that based on the same background data, both versions of explanation data, that is, the original and under-sampled validation set, gave comparable quantitative results of AUC. With *p* increasing to 0.5, the corresponding top-ranked variables tend to have greater predictive power, suggesting our proposed balancing strategy is superior than using background and explanation data with a random minority-overall rate.

Although we only presented results for Deep SHAP in the main analysis, our method also applies for Gradient SHAP. In an additional analysis, we explained the same MLP model trained on our dataset using Gradient SHAP and presented the results in the supplementary material (eFigure 1-3). Gradient SHAP is more time-consuming than Deep SHAP, since it builds many sub-models from scratch within the explainer in order to calculate the variable contributions, whereas Deep SHAP acts in the way of approximation utilizing DeepLIFT. As the background data matters to DeepLIFT[7], it also matters to Deep SHAP, and hence, the data imbalance can have a more dramatic effect on the explanation results than with Gradient SHAP, as shown in eFigure 1. Furthermore, we highlight that our proposed adjustments to Deep SHAP and Gradient SHAP are not specific to the explanation of MLP, but are generally applicable to SHAP explanation of any deep learning models.

In the event of data imbalance, SHAP has been widely applied without adjustment for model explanation in many medical applications[13, 14, 16]. However, our study demonstrates that the issue of data imbalance may lead to unreliable model explanations, just as it does for model predictions. BalanceSHAP is an easy-to-use and lightweight tool designed to assist researchers in finding more plausible explanations in the case of data imbalance.

This Python library, complementary to SHAP, can be useful to visualize additional information on interpretation, such as variable outliers, which is especially valuable in the face of high-dimensional input of black box models.

There are several limitations to this study. First, we focused on structured health data. The effects of data imbalance on model explanation may differ from data types such as image data, which are worth exploring in the future. Second, we only set one optimized model for explanation with the uncertainty for variable importance. Lastly, we only applied beeswarm plot examination and testing ability comparison to evaluate explanation results. Further research may extend to other dimensions, such as SHAP-based partial dependence and pivot plots.

## 5 Conclusion

In this empirical study, we demonstrated that in the presence of the data imbalance, balanced background and explanation data improved SHAP's explanation results in terms of reducing the data imbalance-induced "abnormal points" in the beeswarm plot and enhancing the predictive ability. Our proposed BalanceSHAP can be particularly valuable in the medical field, where data imbalance is a prevalent problem.

**Figure 1** Comparison of beeswarm plots based on (a) unbalanced background data, with a minority-overall rate of 0.088 and (b) balanced background data, with a minority-overall rate of 0.5.

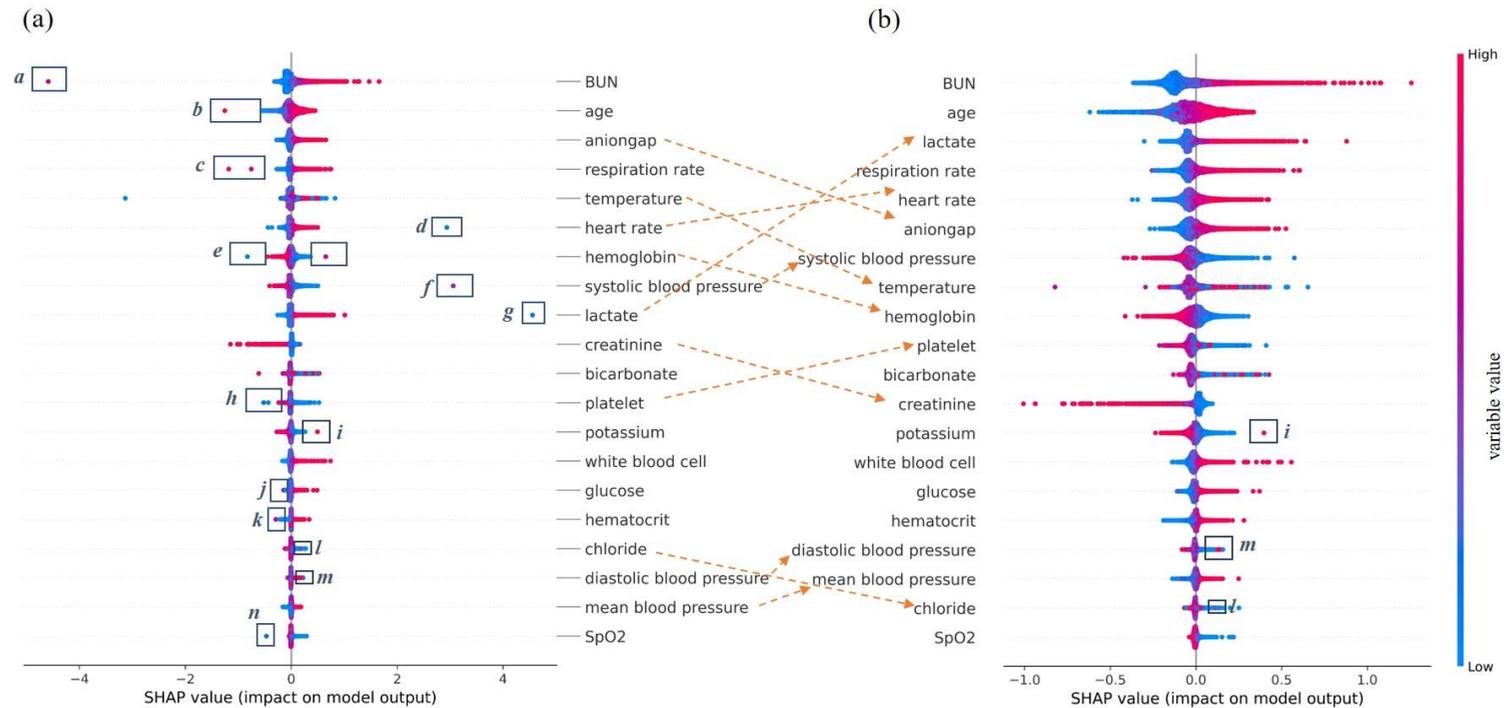

Based on the same explanation data, when the background minority-overall rate was changed to 0.5, ranks for variables anion gap, temperature, hemoglobin, creatinine, and chloride were dropped, while ranks for heart rate, systolic blood pressure, lactate, platelet, diastolic blood pressure, and mean blood pressure were raised. For points *a-n*, the "outliers" of trend in the (a) were amended in (b), except for "outliers" points *i*, *m*, and *l*. BUN, blood urea nitrogen; SpO2, peripheral capillary oxygen saturation.

**Figure 2** Comparison of beeswarm plots based on (a) unbalanced explanation data, which was the original validation set and without under-sampling and (b) balanced explanation data, which was the under-sampled validation set.

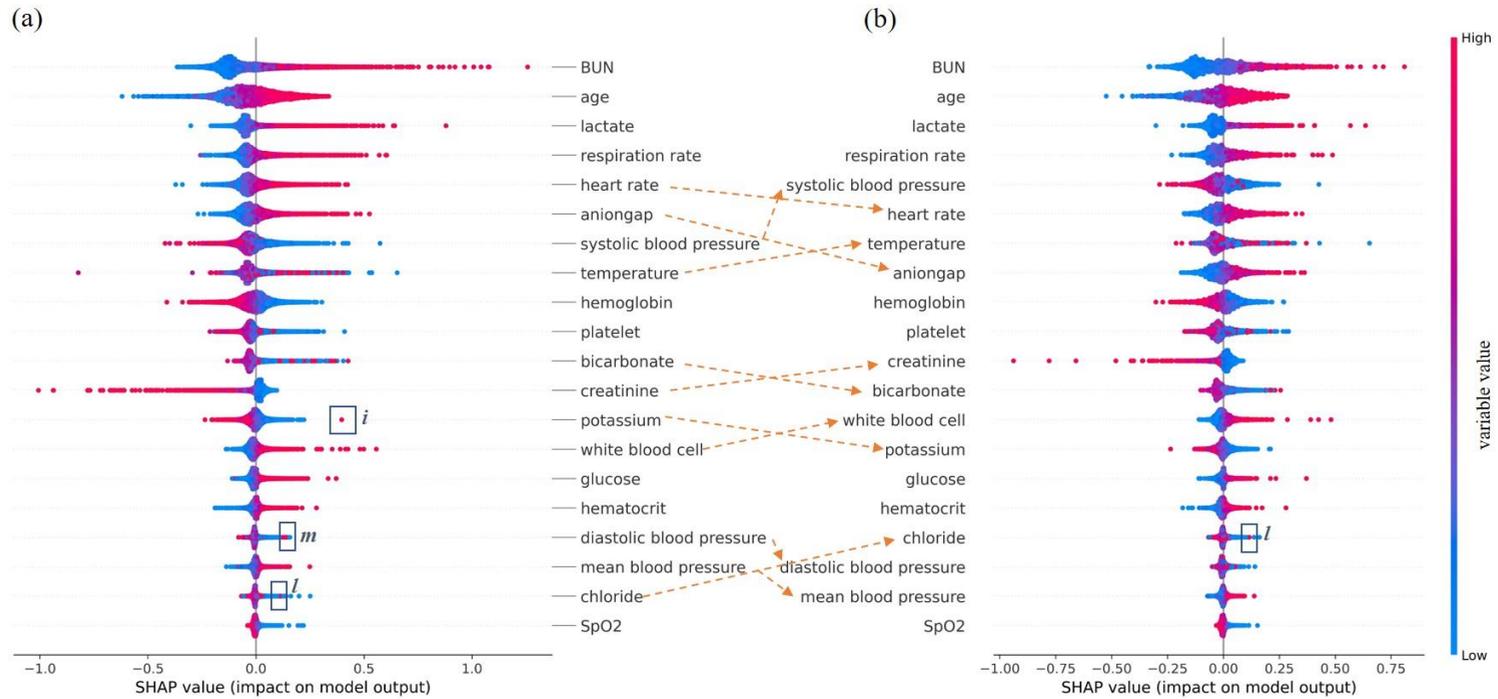

Based on balanced background data i.e. minority-overall rate of 0.5, with a balanced explanation data, ranks for variables heart rate, anion gap, bicarbonate, potassium, diastolic blood pressure, and mean blood pressure were lowered down. In contrast, ranks for systolic blood pressure, temperature, creatinine, white blood cell, and chloride were raised. For points *i* and *m*, the "outlier" in the (a) were amended in the right panel due to the adjustment on explanation data, while "outlier" *l* remains in the plot. The sub-figure (a) here is the same as sub-figure (b) in the Figure 1. BUN, blood urea nitrogen; SpO2, peripheral capillary oxygen saturation.

**Figure 3** Comparison of predictivity performance for MLPs built with top variables yielded by unbalanced background and explanation data (black) and balanced background and explanation data (blue).

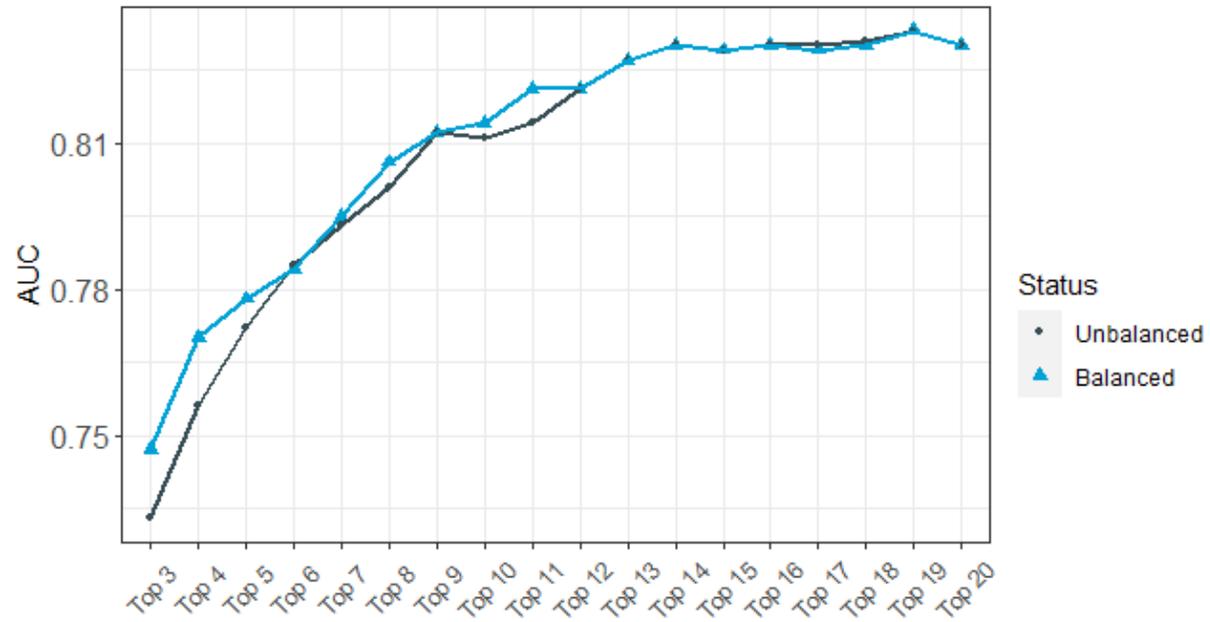

The unbalanced background and explanation data have a minority-overall rate of 0.088, the same as the event rate, and the balanced background and explanation data have a minority-overall rate of 0.5.

# Supplementary Materials

**eTable 1.** AUC for MLP models built by Top $k$ ($k$ = 3, 4, …, 20) variables from importance rankings produced by SHAP values using background minority-overall rates of 0.088, 0.2, 0.3, 0.4, and 0.5, with or without under-sampling for explanation data.

|  | Original | Minority-overall rate = 0.2 | | Minority-overall rate = 0.3 | | Minority-overall rate = 0.4 | | Minority-overall rate = 0.5 | |
| --- | --- | --- | --- | --- | --- | --- | --- | --- | --- |
|  |  | No US | US | No US | US | No US | US | No US | US |
| Top 3 | 0.733 | 0.736 | 0.736 | 0.736 | 0.736 | 0.736 | 0.736 | **0.747** | **0.747** |
| Top 4 | 0.756 | 0.753 | 0.753 | 0.753 | 0.748 | 0.756 | 0.748 | **0.770** | **0.770** |
| Top 5 | 0.772 | 0.770 | 0.763 | 0.766 | 0.763 | 0.766 | 0.763 | **0.778** | 0.775 |
| Top 6 | 0.785 | 0.785 | 0.782 | 0.785 | 0.782 | 0.785 | **0.787** | 0.784 | **0.787** |
| Top 7 | 0.793 | 0.793 | 0.787 | 0.794 | 0.794 | 0.794 | **0.800** | 0.795 | **0.800** |
| Top 8 | 0.801 | 0.801 | 0.801 | 0.801 | 0.801 | **0.806** | **0.806** | **0.806** | **0.806** |
| Top 9 | 0.812 | 0.812 | 0.812 | 0.812 | 0.812 | 0.812 | 0.812 | 0.812 | 0.812 |
| Top 10 | 0.811 | 0.811 | **0.814** | 0.813 | **0.814** | 0.813 | **0.814** | **0.814** | **0.814** |
| Top 11 | 0.814 | 0.814 | **0.821** | 0.814 | 0.820 | **0.821** | 0.820 | **0.821** | 0.820 |
| Top 12 | 0.821 | 0.821 | 0.821 | 0.821 | 0.821 | 0.821 | 0.821 | 0.821 | 0.821 |
| Top 13 | 0.827 | 0.827 | 0.826 | 0.827 | 0.826 | 0.827 | 0.826 | 0.827 | 0.826 |
| Top 14 | 0.830 | 0.830 | 0.830 | 0.830 | 0.830 | 0.830 | 0.830 | 0.830 | 0.830 |
| Top 15 | 0.829 | 0.829 | 0.829 | 0.829 | 0.829 | 0.829 | 0.829 | 0.829 | 0.829 |
| Top 16 | 0.830 | 0.830 | 0.830 | 0.830 | 0.830 | 0.830 | 0.830 | 0.830 | 0.830 |
| Top 17 | 0.830 | 0.830 | 0.830 | 0.830 | 0.830 | 0.830 | 0.830 | 0.829 | 0.830 |
| Top 18 | 0.831 | 0.831 | 0.831 | 0.831 | 0.831 | 0.831 | 0.831 | 0.830 | 0.831 |
| Top 19 | 0.833 | 0.833 | 0.833 | 0.833 | 0.833 | 0.833 | 0.833 | 0.833 | 0.833 |
| Top 20 | 0.830 | 0.830 | 0.830 | 0.830 | 0.830 | 0.830 | 0.830 | 0.830 | 0.830 |

The value(s) in bold is (are) the top values in each row. The shaded cells show the results yielded with balanced background data containing all bold values for Top 3-4 and half of the bold values for Top 5-11. Minority-overall rate:

minority-overall rate on background data; US: under-sampling for explanation data, when the background data was with the minority-overall rate of 0.5, under-sampled version of explanation data was with a minority-overall rate of 0.5, and then both of them could be considered as balanced.

**eFigure 1** With Gradient SHAP, Comparison of beeswarm plots based on (a) unbalanced background data, with a minority-overall rate of 0.088 and (b) balanced background data, with a minority-overall rate of 0.5.

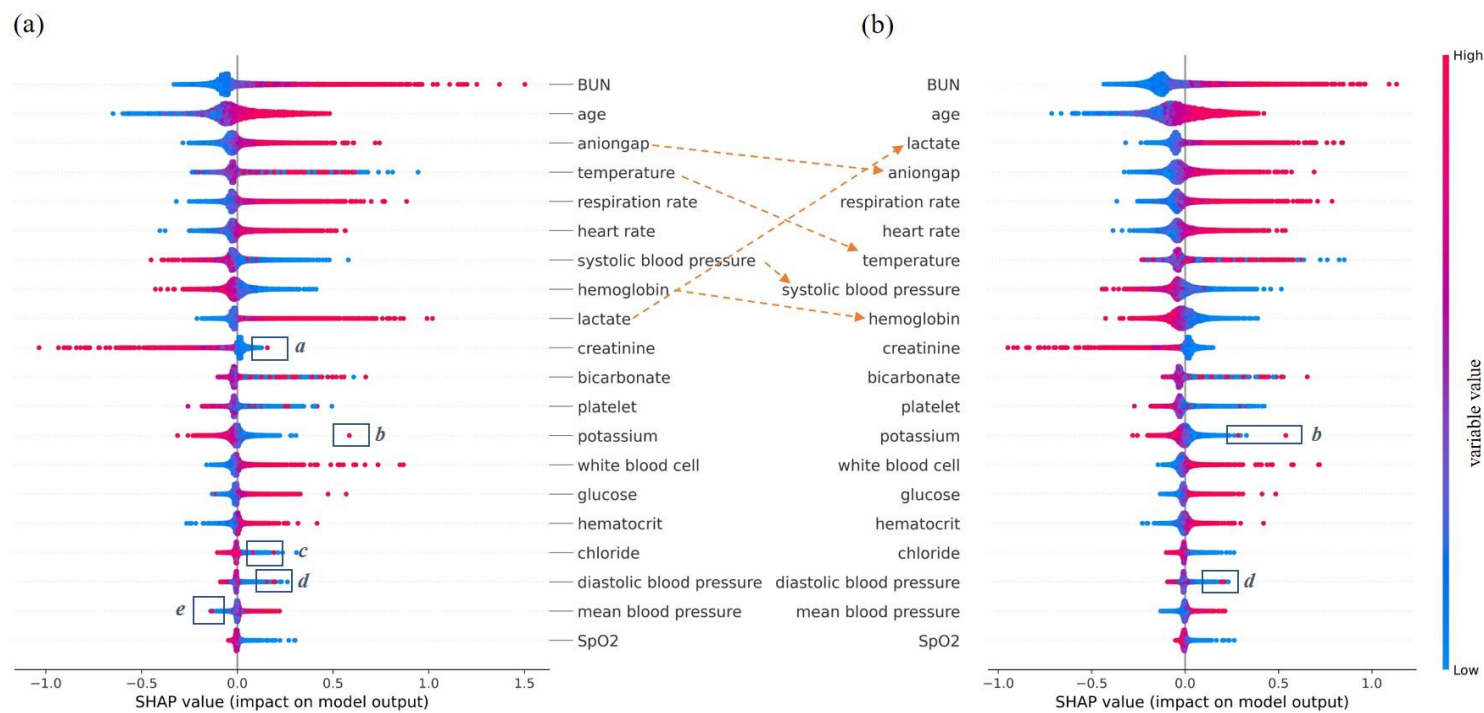

Based on the same explanation data, when the background minority-overall rate was changed to 0.5, ranks for variables anion gap, temperature, and hemoglobin were dropped, while ranks for lactate was raised. For points *a-e*, the "outliers" of trend in the (a) were amended in the (b). The "outliers" remain in the panels for points *b* and *d*. The sub-figure (a) here is the same as sub-figure (b) in the eFigure 1. BUN, blood urea nitrogen; SpO$_2$, peripheral capillary oxygen saturation.

**eFigure 2** With Gradient SHAP, Comparison of beeswarm plots based on (a) unbalanced explanation data, which was the original validation set and without under-sampling and (b) balanced explanation data, which was the under-sampled validation set.

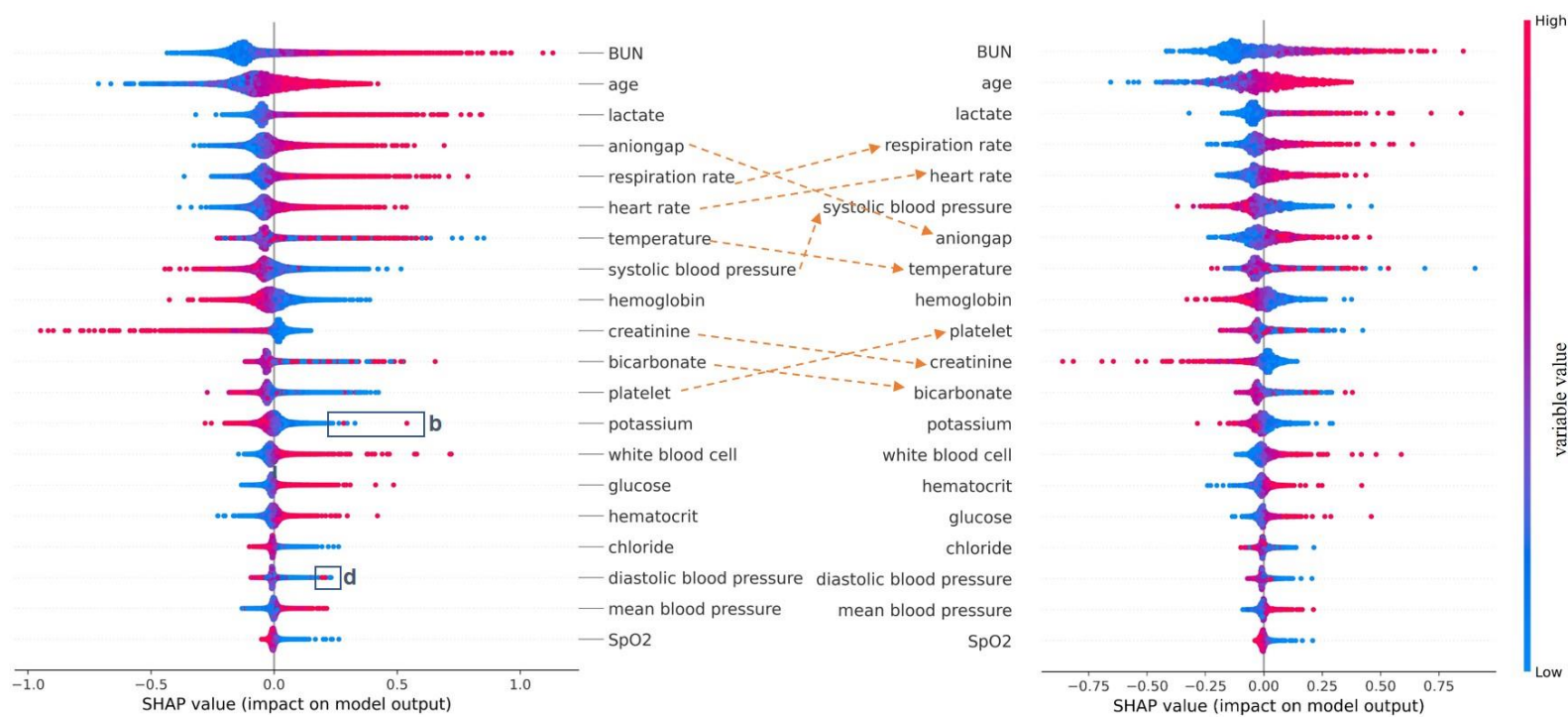

Based on balanced background data i.e. with a minority-overall rate of 0.5, with a balanced explanation data, ranks for variables anion gap, temperature, creatine and bicarbonate were lowered down. In contrast, ranks for respiration rate, heart rate, systolic blood pressure, and platelet were raised. For points *b* and *d*, the "outliers" in (a) were amended in (b) due to the adjustment of explanation data. BUN, blood urea nitrogen; SpO$_2$, peripheral capillary oxygen saturation.

**eFigure 3** With Gradient SHAP, comparison of predictivity performance for MLPs built with top variables yielded by unbalanced background and explanation data (black) and balanced background and explanation data (blue).

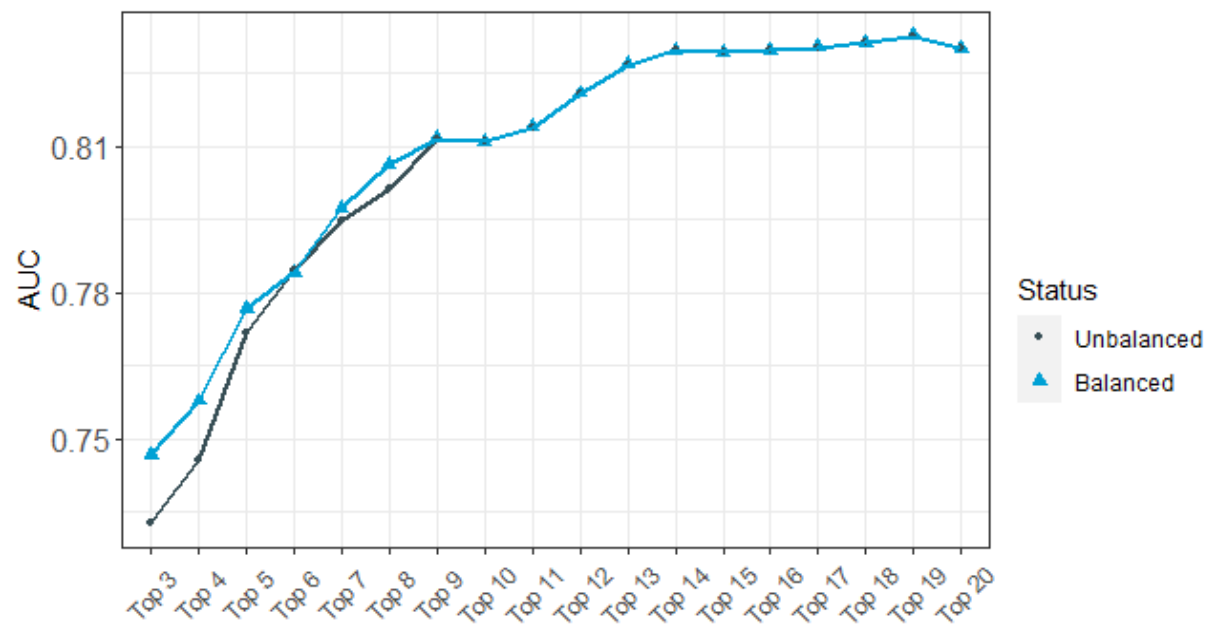

The unbalanced background and explanation data have a minority-overall rate of 0.088, the same as the event rate, and the balanced background and explanation data have a minority-overall rate of 0.5.